\documentclass{article}

\usepackage[nonatbib, preprint]{nips_2018}

\usepackage[utf8]{inputenc} 
\usepackage[T1]{fontenc}    
\usepackage{hyperref}       
\usepackage{url}            
\usepackage{booktabs}       
\usepackage{amsfonts}       
\usepackage{nicefrac}       
\usepackage{microtype}      
\usepackage{amsmath}
\usepackage{amssymb}
\usepackage{commath}
\usepackage{graphicx}
\usepackage{mathtools}
\usepackage{multirow}
\usepackage{float}
\usepackage{wrapfig}
\usepackage[square,numbers]{natbib}
\usepackage{footmisc}
\DefineFNsymbols{mySymbols}{{\ensuremath\dagger}{\ensuremath\ddagger}\S\P
  *{**}{\ensuremath{\dagger\dagger}}{\ensuremath{\ddagger\ddagger}}}
\setfnsymbol{mySymbols}

\title{Learning and Inferring Movement with Deep Generative Model}

\author{
Mingxuan Jing\thanks{These two authors contributed equally} \\
Tsinghua University \\
\And
Xiaojian Ma\footnotemark[1] \\
Tsinghua University \\
\AND
Fuchun Sun \\
Tsinghua University \\
\And
Huaping Liu \\
Tsinghua University\\
}

\begin{document}

\maketitle


\begin{abstract}
    Learning and inference movement is a very challenging problem due to its high dimensionality and dependency to varied environments or tasks. In this paper, we propose an effective probabilistic method for learning and inference of basic movements. The motion planning problem is formulated as learning on a directed graphic model and deep generative model is used to perform learning and inference from demonstrations. An important characteristic of this method is that it flexibly incorporates the task descriptors and context information for long-term planning and it can be combined with dynamic systems for robot control. The experimental validations on robotic approaching path planning tasks show the advantages over the base methods with limited training data.
\end{abstract}

\section{Introduction} 
Motion planning is abstruse in robotics. Traditionally, the mainstream solutions are based on searching \cite{rrt}\cite{PRM} or stochastic optimization \cite{STOMP}\cite{Particle}. However, such algorithms are highly depended on accurate dynamics or known environments, which can be hard to obtain for diverse real-world tasks and makes these algorithms expensive to run. From another perspective, some recent work models this problem as a decision process and utilizes reinforcement learning(RL) to learn a policy for specific robotic task \cite{RL1}\cite{RL2}. But such RL-based methods usually require amounts of explorations that can only be done in simulators(due to the safety issues) \cite{saferl}, and the learned policy itself can also be unstable \cite{RLMATTERS}. In general, effective and stable motion planning still remains an open problem in robotics.
    

It's wildly expected that a successful long-term motion planning will be composed of many desired sub-plannings \cite{planning2}\cite{planning1}. Thus the purpose is to build a model that can learn and inference relatively basic movements, i.e., approaching, hitting, grasping, etc, and the model should also be able to take into new commands and environment-rated information for continuous movement generation. Most importantly, essential mechanisms are needed to guarantee the stabilization and safety, both in learning and inference phase. 

In this paper, we propose a motion planning framework named Contextual Movement Primitives (CMP). CMP aims at effective learning and inference of basic movements, while provides the interface for commands(tasks) and environment(context) information as input for long-term planning. By introducing the probabilistic representation of trajectory, we model the motion planning problem as learning on a directed graphic model. Though the probabilistic distribution can be really complicated, with the help of the deep generative model, we make it possible to perform learning and inference on this probabilistic motion planning model from demonstrations. Finally, we extend the probabilistic trajectory representation to force space and show an elegant combination with a dynamic control system. The dynamic control system does bring us lots of desired properties on controlling a real robot, and significantly improve the qualities of generated movements.
    
To summarize, our main contributions can be listed as follows:
\begin{itemize}
    \item We propose Context Movement Primitives(CMP), a probabilistic trajectory representation and learning framework based on a deep generative model. CMP is able to handle high-dimension conditional input including context information and task description, and generate feasible and practical trajectories for controlling a real robot under real-world tasks.
    \item We combine CMP with a dynamic control system, empower it the ability to directly learn the control value of a dynamic system, while the internal properties of the dynamic system like robust to perturbation and goal-convergence are maintained.
    \item We achieve competitive performance on approaching path planning task both in simulation and real robot than supervised baselines \cite{CNNDMP} with limited data, which shows that CMP has the potential to be a general framework for robot trajectory and skill learning.
\end{itemize}

\section{Preliminaries} 
In this section, several conceptions would be introduced for facilitating further discussions.

\subsection{Probabilistic Trajectory Representation}\label{Probabilistic Trajectory Representation}
Generally, a certain trajectory $\boldsymbol{\tau}=f(\mathbf{t}): \mathbf{R} \mapsto \mathbf{R^n}$ could be denoted as a time-varying function or sequential-time sampling over that function. For the flexibility on time re-scaling, a monotonous increased phase variable $\mathbf{s} \coloneqq \mathbf{s(t)}: \mathbf{R} \mapsto [0, 1]$ could be used instead of actual execution time $\mathbf{t}$, so that the trajectory could be rewritten as $\boldsymbol{\tau}=f(\mathbf{s})$.

As described above, $f(\mathbf{s})$ could be any continuous function, which is catastrophic for learning directly. For a smooth trajectory whose values at each time point are closely co-related with the value in nearby time phase, a reasonable method may be approximating $f(\mathbf{s})$ by limited and fixed number of parameters using Radial Basis Function(RBF) approximater. RBF uses kernel functions $\boldsymbol{\Phi}^T(\mathbf{s}) = [\boldsymbol{\phi}_1(\mathbf{s}), \boldsymbol{\phi}_2(\mathbf{s}), \cdots]$ as basic components and regard mixture weights $\mathbf{w}$ as parameters. $ \boldsymbol{\tau} = f(\mathbf{s}) = \boldsymbol{\Phi}^T(\mathbf{s}) \mathbf{w} + \boldsymbol{\epsilon}_\tau$, where $\boldsymbol{\epsilon}_\tau \sim N(0, \boldsymbol{\Sigma}_\tau)$ denotes approximation residuals.

When modeling the property of trajectory variability under perturbations \cite{ruckert2013learned} on certain environment, a trajectory could be seen as composed by deterministic kernel functions at each time phase but with stochastic mixture weight $\mathbf{w}$ drawn from a distribution. For most common case, Gaussian distribution is used here: $\mathbf{w} \sim N(\boldsymbol{\mu}_w, \boldsymbol{\Sigma}_w)$, and the probability of a trajectory $\boldsymbol{\tau}$ could be calculated as:
\begin{align}\label{trajrecovery}
p(\boldsymbol{\tau}) = \prod_s p(f(\mathbf{s})|\mathbf{w}) p(\mathbf{w}) = \prod_s N\left(f(\mathbf{s})|\boldsymbol{\Phi}^T(\mathbf{s}) \mathbf{w}, \boldsymbol{\Sigma}_\tau\right) N\left(\mathbf{w}|\boldsymbol{\mu}_w, \boldsymbol{\Sigma}_w\right)
\end{align}
As described in \cite{paraschos2013probabilistic}, here $\boldsymbol{\Sigma}_\tau$ is a fixed value once weight $\mathbf{w}_\tau$ is calculated with given $\boldsymbol{\Phi}(\mathbf{s})$ on $\boldsymbol{\tau}$. The optimal $\mathbf{w}_\tau$ can be calculated by maximum logistic likelihood estimation\footnote[1]{Since the trajectory variance $\boldsymbol{\Sigma}_w$ is fixed, and $N(\boldsymbol{\mu}_w, \boldsymbol{\Sigma}_w)$ is isotropic, \eqref{ori l2} could then be rewritten as \eqref{end l2}.}:
\begin{align}
    \max_w \log p(\boldsymbol{\tau}|\mathbf{w}) &= \max_{w, \Sigma_\tau} \sum_s \log\frac{\exp\left(-\frac{1}{2}(\boldsymbol{\tau} - \boldsymbol{\Phi}^T(\mathbf{s}) \mathbf{w})^T\boldsymbol{\Sigma}_\tau^{-1}(\boldsymbol{\tau} -\boldsymbol{\Phi}^T(\mathbf{s}) \mathbf{w})\right)}{|\boldsymbol{\Sigma}_\tau|\sqrt{(2\pi)^{dim(\boldsymbol{\tau})}}}\label{ori l2}\\
    \Rightarrow \max_w \log p(\boldsymbol{\tau}|\mathbf{w}) &= \min \sum_s \norm{\boldsymbol{\tau} - \boldsymbol{\Phi}^T(\mathbf{s}) \mathbf{w}}_{2}\label{end l2}
\end{align}
Now a trajectory $\boldsymbol{\tau}$ can be regarded as a sample draw from distribution shown in \eqref{trajrecovery} with parameter $\boldsymbol{\mu}_w$ and $\boldsymbol{\Sigma}_w$, which can be learned and inferred under probabilistic model. Given $\mathbf{w} \sim N(\boldsymbol{\mu}_w, \boldsymbol{\Sigma}_w)$, a trajectory can be generated by computing $E_{\boldsymbol{\tau}\sim p(\boldsymbol{\tau} | \mathbf{w})}[\boldsymbol{\tau}] = \boldsymbol{\Phi^T(s)\mathbf{w}}$. In the rest of this paper, we will consider $\mathbf{w}$ and its distribution $p(\mathbf{w})$ as the \textbf{trajectory probabilistic representation}.

For trajectory learning, usually, we are more interested in whether a trajectory can be generated with some conditions $\mathbf{c}$ that related to a specific task. This is exactly what we concentrate on in this paper. Thus we will discuss the details about learning $p(\mathbf{w}|\mathbf{c})$ instead of $p(\mathbf{w})$ in the following.

\subsection{Dynamic Control System}\label{Dynamic Control System}
A trajectory generator or robot movement controller should have the ability to handle the dynamic properties and perturbations when applied on real-world robots. Overall, a control system should hold these characteristics: 1) keeping a trajectory dynamically feasible, so that the trajectory can be executed on real-world; 2) guaranteeing to reach target point at the end of the trajectory, so that the task could be accomplished; 3) being robust on noise or robot control errors so that the performance could be reliable; 4) the control signal needs to be efficiently learned and generated. 

A well designed second-order dynamic control system with force signal $f_{ctl}(\mathbf{s}): \lim_{\mathbf{s} \to 1}f_{ctl}(\mathbf{s}) = 0$ and end constraint $\mathbf{g}$, would properly satisfy these required properties and generate trajectories by solving the system dynamic numerically:
\begin{align}\label{DMP_def}
    \boldsymbol{\ddot{\tau}} &= \mathbf{K}(\mathbf{g} - \boldsymbol{\tau}) - \mathbf{B}\boldsymbol{\dot{\tau}} + f_{ctl}(\mathbf{s}; \mathbf{w})
\end{align}
 $\mathbf{K}$, $\mathbf{B}$ are constant system parameters that control the time domain attributes of this dynamic system. This kind of dynamic system is analytically well understood that holds stability properties, which means when integrated, it would generates the desired behavior under a given $f_{ctl}$ and $\mathbf{g}$. 

An interesting fact is, here $f_{ctl}(\mathbf{s})$ could be regarded as a force space trajectory and thus we can describe it with the same approach as a normal trajectory(mentioned in Section \ref{Probabilistic Trajectory Representation}). This means that it's possible to alternatively represent and learn the force space trajectory instead of the common trajectory with position,  then control the robot through a dynamic control system. As we discussed above, by learning force space trajectories, we can obtain some control-friendly properties brought by the dynamic control system \cite{pastor2011skill}. More relevant details will be discussed in the following.

\section{Problem Formalization}\label{Problem Formalization}

\begin{wrapfigure}{l}{0.35\textwidth}
\begin{center}
    \includegraphics[width=1.0\linewidth]{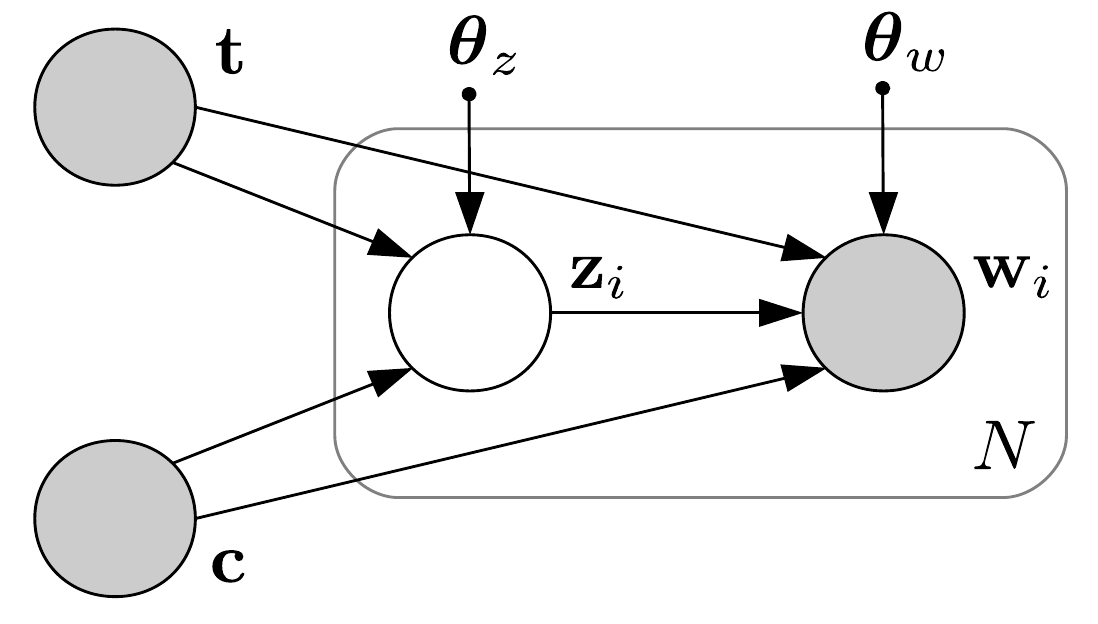}
\end{center}
\vskip -0.1 in
   \caption{Graphical model underlying our approach. Shaded nodes are all observed during learning, and our model learns parameters $\boldsymbol{\theta} = \{\boldsymbol{\theta}_z, \boldsymbol{\theta}_w\}$. During testing, only $\mathbf{c}$, $\mathbf{t}$ are available.}
\label{graphical_model}
\vskip -0.1 in
\end{wrapfigure}

We denotes an individual trajectory representation as $\mathbf{w}$, contextual information $\mathbf{c}$, and a descriptor to the given task as $\mathbf{t}$. We introduce a set of continuous latent variable $\mathbf{z}$ with prior distribution $p_{\boldsymbol{\theta}_z}(\mathbf{z}|\mathbf{c}, \mathbf{t})$. Then we can define the conditional distribution of $w$ as $p_{\boldsymbol{\theta}_w}(\mathbf{w}|\mathbf{c},\mathbf{t},\mathbf{z})$. A visual illustration of the underlying graphical model is shown in Figure \ref{graphical_model}. Given contextual information and task taken from demonstrations $D = \{(\mathbf{w}_1, \mathbf{c}_1, \mathbf{t}_1), (\mathbf{w}_2, \mathbf{c}_2, \mathbf{t}_2) ... (\mathbf{w}_n, \mathbf{c}_n, \mathbf{t}_n)\}$, the problem will be learning on this directed probabilistic model with maximum likelihood, in presence of continuous latent variables of $p_{\boldsymbol{\theta}_z}$ with intractable posterior distribution $p_{\boldsymbol{\theta}_w}$. $\boldsymbol{\theta} = \{\boldsymbol{\theta}_z, \boldsymbol{\theta}_w\}$ will be the parameters of the expected generative model. When the model is learned, a feasible trajectory under specified context and task can be inferred on it. 

\textbf{Representation of Contextual Information and Task}~~~In our framework, $\mathbf{c}$ can be varied. But here we simply provide the RGB photograph taken from the First-Person view of the working scene, where the trajectory will be executed, as contextual information. As for the task descriptor, we assume that there exists a set $\mathbb{T}$ with finite tasks. Then $\mathbf{t}$ is a one-hot tensor of size $\abs{\mathbb{T}}$. Each bit in $\mathbf{t}$ is associated with a single task. Thus we can see $\mathbf{t}$ as a random variable derived from $\mathbf{t} \sim Multinomial(n=1, p_0 = ... = p_k = 1/k)$. 

\section{Contextual Movement Primitives} 

\subsection{Model Overview}

\begin{figure}[!ht]
\begin{center}
    \includegraphics[width=1.0\linewidth]{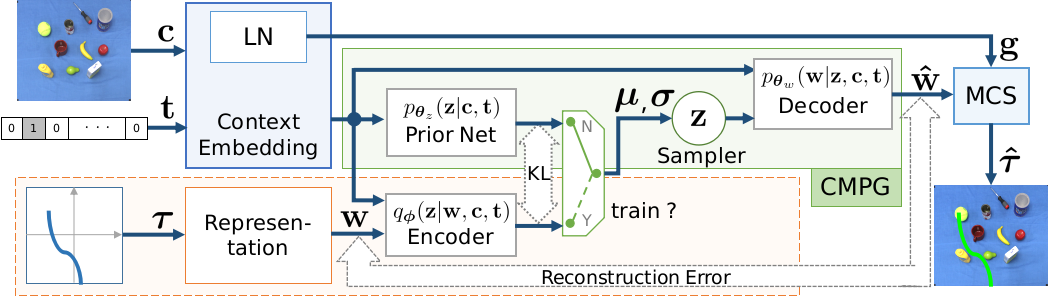}
\end{center}
\vskip -0.1 in
   \caption{Illustration of our model. Besides CMPG, LN and MCS, an encoder is needed during training. The inner structure of Context Embedding and LN can be found in Figure \ref{LN}.}
\label{flowchart}
\vskip -0.1 in
\end{figure}

In order to achieve good performance on trajectory learning problems, we propose Contextual Movement Primitives(CMP), a probabilistic trajectory representation and learning framework that effectively generates trajectories for specific context and task in an end-to-end manner. By combining CMP with a dynamic control system, a robot can operate under the control value output by our framework directly, and acquire some properties of dynamic control system like perturbation-robust and goal-convergence, which are extremely important for robots and tasks in real-world.

Our CMP framework consists of 3 modules: CMP generator(CMPG), localization network(LN) and motion control system(MCS). An overview of it is shown in Figure \ref{flowchart}. All the modules composite CMP together to enable end-to-end movement learning and inference. We will show the details of each module below, and provide the full specification in the Appendix.

\subsection{Movement Inference and Trajectory Generation}
In our framework, CMP generator(CMPG) serves as a generative model that receives the contextual information $\mathbf{c}$ and task descriptor $\mathbf{t}$ as input, then performs an inference on the learned graphical model about $p_{\boldsymbol{\theta}_w}(\mathbf{w}|\mathbf{c}, \mathbf{t}, \mathbf{z})$ to produce a representation $\mathbf{w}$ of trajectory. However, learning on this graphical model can be challenging, since latent variable $\mathbf{z}$ is continuous and $p_{\boldsymbol{\theta}}$ is intractable. We solve this issue following Welling~\textit{et.al.}~\cite{kingma2014auto} and Lee~\textit{et.al.}~\cite{sohn2015learning} by maximizing the \textit{evidence lower bound}(ELBO) $\mathcal{L}(\boldsymbol{\theta}, \boldsymbol{\phi}, \mathbf{w}_i, \mathbf{c}_i, \mathbf{t}_i)$ of the likelihood over demonstrations $D$(Section \ref{Problem Formalization}):
\begin{align}
    \log p_{\boldsymbol{\theta}}(\mathbf{w}_i | \mathbf{c}_i, \mathbf{t}_i) \geq \mathcal{L}(\boldsymbol{\theta}, \boldsymbol{\phi}, \mathbf{w}_i, \mathbf{c}_i, \mathbf{t}_i) =\mathbb{E}_q\left[ -\log q_{\boldsymbol{\phi}}(\mathbf{z}|\mathbf{w}_i, \mathbf{c}_i, \mathbf{t}_i) + \log p_{\boldsymbol{\theta}_w}(\mathbf{w}_i|\mathbf{z}, \mathbf{c}_i, \mathbf{t}_i)\right]
\end{align}
which can also be written as:
\begin{align}\label{ELBO}
    \mathcal{L}(\boldsymbol{\theta}, \boldsymbol{\phi}, \mathbf{w}_i, \mathbf{c}_i, \mathbf{t}_i) = - \mathit{KL}\left(q_{\boldsymbol{\phi}}(\mathbf{z}|\mathbf{w}_i, \mathbf{c}_i, \mathbf{t}_i) || p_{\boldsymbol{\theta}_z}(\mathbf{z} | \mathbf{c}_i, \mathbf{t}_i) \right) + \mathbb{E}_q\left[\log p_{\boldsymbol{\theta}_w}(\mathbf{w}_i|\mathbf{z}, \mathbf{c}_i, \mathbf{t}_i)\right]
\end{align}
Notice that the lower bound above is slightly different from~\cite{kingma2014auto}. It is extended to a conditional distribution. We provides an mechanism closed to~\cite{sohn2015learning} under encoder-decoder structure to differentiate and optimize the conditional lower bound $\mathcal{L}(\boldsymbol{\theta}, \boldsymbol{\phi}, \mathbf{w}_i, \mathbf{c}_i, \mathbf{t}_i)$ w.r.t. both the variational parameters $\boldsymbol{\phi}$ and generative parameters $\boldsymbol{\theta}$. The model is illustrated in Figure \ref{flowchart} with an encoder, a decoder and a prior network. Encoder is a neural network that maps $\mathbf{c}$, $\mathbf{t}$ and $\mathbf{w}$ into latent space $\mathbf{z}_i \sim q_{\boldsymbol{\phi}}(\mathbf{z}|\mathbf{w}_i, \mathbf{c}_i, \mathbf{t}_i)$ with parameters $(\boldsymbol{\mu}_e(\mathbf{w}_i, \mathbf{c}_i, \mathbf{t}_i), \boldsymbol{\Sigma}_e(\mathbf{w}_i, \mathbf{c}_i, \mathbf{t}_i))$. While decoder take the input of $\mathbf{z}_i$ from encoder to approximate the posterior $p_{\boldsymbol{\theta}_w}$. The first KL-divergence term in \eqref{ELBO} measure how close the distribution $q_{\boldsymbol{\phi}}$ to $p_{\boldsymbol{\theta}_z}$, while the later one will be represented by a neural network named prior network. Thus the KL term can be computed in closed-form. The second term of expectation over $q_{\boldsymbol{\phi}}$ can be formalized by concatenating the encoder and decoder, then minimizing the reconstruction error.

During training, reparameterization trick \cite{kingma2014auto} will be used to re-write the expectation w.r.t. $q_{\boldsymbol{\phi}}$ in \eqref{ELBO}. In each iteration, the encoder will generate the parameter $(\boldsymbol{\mu}_e, \boldsymbol{\Sigma}_e)$ of $q_{\boldsymbol{\phi}}$, then a reparameterization of latent variable $\mathbf{z} = \boldsymbol{\mu}_e + \boldsymbol{\Sigma}_e\boldsymbol{\epsilon}, ~~\boldsymbol{\epsilon} \sim N(\mathbf{0}, \mathbf{I})$ will be fed into decoder. Such that the Monte Carlo estimation of the second term in \eqref{ELBO} is differentiable w.r.t. $\boldsymbol{\phi}$. When all the parameters are learned, the decoder will be reserved as CMPG, it will generate a trajectory representation $\mathbf{w}$ with contextual information $\mathbf{c}$, task descriptor $\mathbf{t}$ and latent $\mathbf{z} \sim p_{\boldsymbol{\theta}_z}(\mathbf{z} | \mathbf{c}_i, \mathbf{t}_i)$. This representation $\mathbf{w}$ will be fed into MCS later to recover the real trajectory or control value for operating the robot. 

Since encoder, decoder and prior network will all take $\mathbf{c}$ and $\mathbf{t}$ as input, a two-stream structure \cite{alexnet} that designed for handling these data is needed. As illustrated in Figure \ref{flowchart}, this two-stream structure is placed before the all the 3 modules. One stream is a convolutional neural network(CNN) that maps the RGB image $\mathbf{c}$ of working scene to an embedding that will be used for predicting the distribution over the latent variable $\mathbf{z}$. For another stream handling task descriptor $\mathbf{t}$, we will just concatenate it with the embedding of $\mathbf{c}$. All the parameters between these two-stream structures are shared.

\textbf{Attention with Spatial Arg-Softmax}~~~Prior work on learning visual-motor policy~\cite{end2endvisual} showed that, Spatial Arg-Softmax on the feature map of the last CNN layer will be helpful for learning useful representation about object position, which can be particularly important in robotics domains~\cite{end2endvisualAE}\cite{end2endvisual}\cite{oneshotvisual}. Spatial Arg-Softmax is a operation that converts the channel-wise feature $\mathbf{a}_{cij}$ into probabilistic distribution $\mathbf{s}_{cij}$, then extract the expected position $(\mathbf{f}_{cx}, \mathbf{f}_{cy})$ in coordinate representation:
\begin{align}\label{spatialargsoftmax}
    \mathbf{s}_{cij} = \exp{\mathbf{a}_{cij}}/\sum_{i'j'}\exp{\mathbf{a}_{ci'j'}} && \mathbf{f}_{cx} = \sum_{ij}\mathbf{s}_{cij}\mathbf{x}_{ij} && \mathbf{f}_{cy} = \sum_{ij}\mathbf{s}_{cij}\mathbf{y}_{ij}
\end{align}
We consider it as an attention mechanism for trajectory learning. Following prior work~\cite{end2endvisual}, and in another implementation of CMPG, we use these feature points $(\mathbf{f}_{cx}, \mathbf{f}_{cy})$ extracted by Spatial Arg-Softmax instead of the origin feature as the embedding of $\mathbf{c}$. The experiments show that this extra operation can significantly improve the qualities of generated trajectories.

\subsection{Target Localization for Position Constraints}
\begin{wrapfigure}{r}{0.4\textwidth}
    \includegraphics[width=1.0\linewidth]{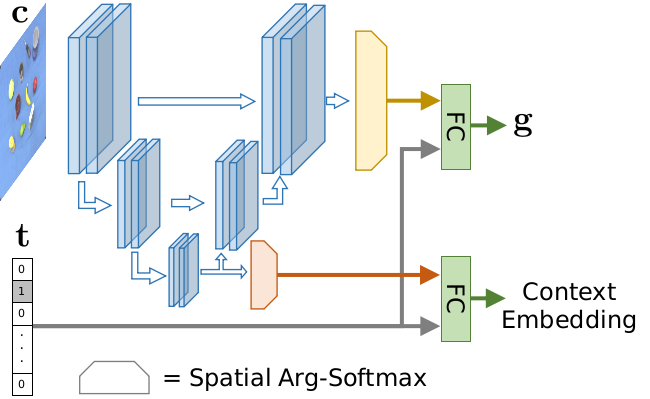}
\vskip -0.1 in
   \caption{The inner structure of Context Embedding and LN. Some layers are shared between these 2 models.}
\label{LN}
\vskip -0.1 in
\end{wrapfigure}
Most of the tasks in robotics domains are target-centric \cite{visualforesight}\cite{zero-shot}\cite{thor}, especially for trajectory learning tasks. This means that accurate position information of task-related objects can be extremely important. Besides the implicit mechanism(Spatial Arg-Softmax) in CMPG for learning position-aware representation, we also introduce localization network(LN) to explicitly predict the position information $d$ of object-of-interest in the current task. Such information will be used as the goal constraints $\mathbf{g}$ in MCS. 

In general, LN will take contextual information $\mathbf{c}$(need to be an image) with task descriptor $\mathbf{t}$ as input, and predict an objectness map. Notice that here we suppose there will be only one object-of-interest for each trajectory generation procedure, thus the regressed position $\mathbf{g} = (\mathbf{f}_{objx}, \mathbf{f}_{objy})$ could then be obtained by performing a Spatial Arg-Softmax over the objectness map. In our implementation, LN reuses some convolutional layers with the two-stream structure in CMPG for handling contextual information. Figure \ref{LN} illustrates the details of LN and feature extractor in CMPG. The training of LN is quite similar to other saliency detection models~\cite{saliency1}\cite{saliency2}. But here we only provide a bounding box of object-of-interest as the ground truth saliency mask. 
\subsection{Trajectory Execution Under Dynamic Control System}
 Supposed that the representation of trajectory $\mathbf{w}$ and target object position(or namely, goal) $\mathbf{g}$ are ready. But we still cannot directly execute them on a robot. MCS is proposed to control the real system with these pieces of information. In our framework, MCS can operate under 2 modes. The main differences between them include how a dynamic control system mentioned in Section \ref{Dynamic Control System} integrates into the trajectory execution and the type of trajectory that $\mathbf{w}$ represents. 

\textbf{MCS as a trajectory solver}~~~For some relatively easy trajectory learning task, a straight-forward approach is directly representing the trajectory of position. In this case, MCS is simply identical as performing a trajectory recovery with $\boldsymbol{\tau} = \mathbb{E}_{p(\boldsymbol{\tau}|\mathbf{w})}[\mathbf{\tau}] = \boldsymbol{\Phi}^T(\mathbf{s})\mathbf{w}, \mathbf{w} \sim N(\boldsymbol{\mu}_w, \boldsymbol{\Sigma}_w)$ as shown in Section \ref{Probabilistic Trajectory Representation}. Once the $\mathbf{w}$ is generated, the corresponding trajectory $\boldsymbol{\tau}$ can be calculated, any position control methods can be applied to control a robot that follows the calculated trajectory. 

\textbf{MCS as a dynamic control system}~~~Although a direct representation of a trajectory of position can be convenient, this approach will suffer from some problems in real-world robot controlling. For example, the shape and end point of a generated trajectory is treated equally in direct representation, but for manipulation tasks, the convergence of generated trajectory to a task-specified goal should be guaranteed. Also, when robot failed to follow the trajectory or a burst perturbation occurs, a control policy should be designed for following the original executing path while keeping the trajectory smooth. We solve these issues by combining CMP with a dynamic control system mentioned in Section \ref{Dynamic Control System}, which will be integrated into CMP as MCS. To control a robot with this version of MCS, we first need to compute the probabilistic representation of force space trajectory $\mathbf{w}$ from a demonstrated trajectory of position $\boldsymbol{\tau}$, which is equal to solving the optimization problem below\footnote[1]{This optimization problem is directly derived from \eqref{DMP_def}.}:
\begin{align}\label{dmpsystem}
    \min_w \norm{\boldsymbol{\Phi}^T(\mathbf{s})\mathbf{w} - \boldsymbol{\ddot{\tau}} - \mathbf{K}(\mathbf{g} - \boldsymbol{\tau}) + \mathbf{B}\boldsymbol{\dot{\tau}}}_{2}
\end{align}
where $\mathbf{g}$ is the goal of current trajectory(for simplification, we use the last point of $\boldsymbol{\tau}$). Training this version of CMPG(with force space trajectory) is basically the same as CMPG with trajectory of position. After the model is converged, force trajectory representation $\mathbf{w}$ can be generated by CMPG after $\mathbf{c}$ and $\mathbf{t}$ are provided. The goal constraints $\mathbf{g}$ can also be obtained from LN simultaneously. 

Now we can control our robot with system dynamics described in \eqref{DMP_def}. For each phase step $\mathbf{s}_i$, we can compute the desired position of a robot in the next step by performing the numerical integration over the system dynamics \eqref{DMP_def} with generated force trajectory representation $\mathbf{w}$. This operation can be performed iteratively until we accomplish all the movements in current task and reach the goal $\mathbf{g}$. Even when a movement is failed to execute properly, i.e. perturbed by others, the execution in next step will not be affected, and the goal constraint can still be guaranteed. More details about the properties of a dynamic control system and the proof of goal convergence can be seen in Appendix.

\section{Experiments} 
We conduct experiments with the Approaching Path Planning task, which will be described in the following. These experiments are designed for answering the questions below:
\begin{itemize}
    \item Can our framework effectively represent the trajectory and learn the posterior of it with contextual information and task?
    \item How does the explicit approach(LN) compare to the implicit mechanism(Spatial Arg-Softmax) on position extraction for learning trajectory under target-centric task?
    \item How does combining dynamic control system(MCS as a dynamic control system) compare to recovering trajectory directly(MCS as a trajectory solver)?
\end{itemize}
For the first question, we prepare environments that cover different domains, including a simulated environment base on Gazebo \cite{gazebo} simulator and an environment in real-world. We will compare the performance of our CMP framework against other baselines on learning and generalization with limited trajectory demonstrations in provided environments. As for the demonstrations $D$, we first synthesis the working scene by randomly selecting and placing the objects, then the trajectories are generated by traditional planners \cite{rrt}\cite{PRM} or simply specified by a human. To make the trajectories more similar to biological movements, zero-mean Gaussian noise with cosine value over the phase will be added. For the latter two questions, we will do ablation analysis by temporarily removing some of the modules in our CMP framework, then test the modified models on a real robot.

\subsection{Approaching Path Planning Task}
\begin{figure}[!ht]
\begin{center}
    \includegraphics[width=1.0\linewidth]{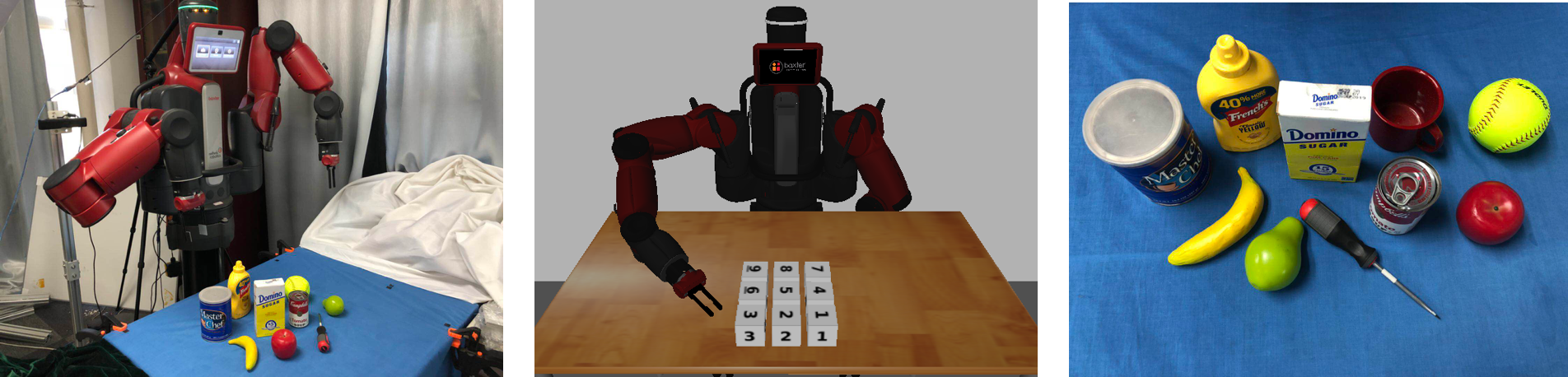}
\end{center}
\vskip -0.1 in
   \caption{Experiment setup. Left: In real-world. Middle: In simulation. Right: Selected object set.}
\label{expsetup}
\vskip -0.1 in 
\end{figure}
We first introduce a set of tasks that will be used in our experiments: \textbf{Approaching Path Planning}. For each task, we need to control a 7-DOF Baxter manipulator to approach a specified target among a cluster of novel objects, pick it up, then back to the initial position. For simulation, we use nine cubes with number 1-9 on them. For the real-world task, all the objects are selected from a subset of YCB object and model set~\cite{ycb}. Experiment setup for simulation/real world and the subset of YCB are shown in Figure \ref{expsetup}. The generalization to non-seen task is non-trivial due to the arbitrary amount and position. This will require the learned model to adapt to totally different contextual information(a different working scene with different objects) and task(different specified target). Furthermore, compared to decision-process based methods~\cite{one-shot}, our model need to generate a complete trajectory, which is even harder due to the length and the smooth requirements of a feasible trajectory. 

\subsection{Trajectory Representation, Learning, and Inference}
In this experiment, we trained the proposed model(CMP) and the baseline deep neural networks(model specifications are listed in the Appendix) on a synthesized demonstrations set. Then test them on another demonstration set with the same size as the training set. We compare the estimations of conditional likelihoods(CLLs) on the testing set of 2 different environments(Simulation and YCB). The early stopping is used during the training based on the estimation of CLLs on the validation set.

\begin{table}
\centering
\begin{tabular}{l||c|c|c|c|c|c} 
\hline
\multirow{2}{*}{CLL} & \multicolumn{2}{c|}{\abs{D}=1000} & \multicolumn{2}{c|}{\abs{D}=2000}  & \multicolumn{2}{c}{\abs{D}=5000}  \\ 
\cline{2-2}\cline{3-7}
& Simulation  & YCB & Simulation  & YCB & Simulation& YCB \\ 
\hline
NN(baseline)  & 1.582   & 4.487    & 2.074   & 4.468   & 2.242 & 5.634  \\
CMP($\mathbf{z} \sim N(\mathbf{0}, \mathbf{I}))$ & 1.674  & 4.460 & 2.093  & \textbf{4.634}   & 3.866 & 5.527  \\
CMP($\mathbf{z} \sim p_{\boldsymbol{\theta}}(\mathbf{c}, \mathbf{t}))$ & \textbf{1.704}  & \textbf{4.490}  & \textbf{2.149}  & 4.581  & \textbf{5.434} & \textbf{5.634}  \\
\hline
\end{tabular}
\vskip 0.1 in
   \caption{The CLLs on synthesized dataset with different sizes.}
\label{exp1table}
\vskip -0.3 in 
\end{table}

\begin{figure}[!ht]
\begin{center}
    \includegraphics[width=1.0\linewidth]{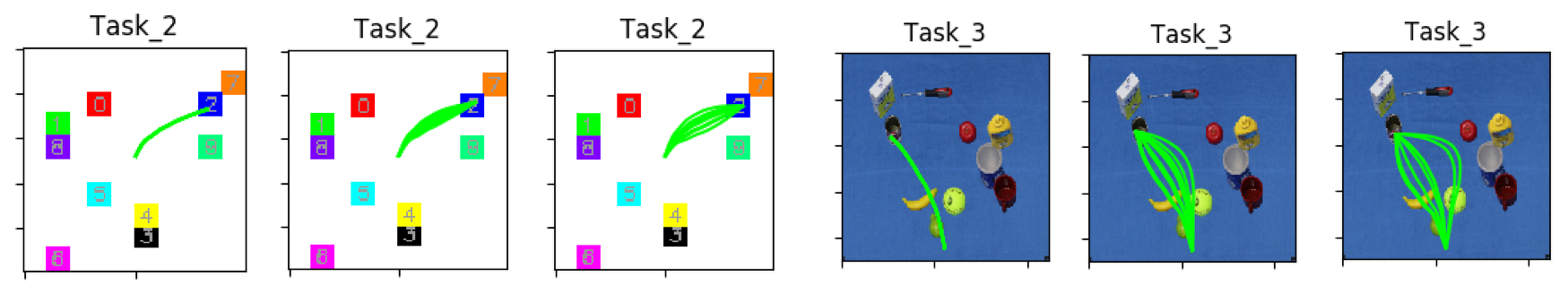}
\end{center}
\vskip -0.2 in
   \caption{Visualization of generated trajectories and working scene(in synthesized simulation/YCB dataset). For each dataset, left is NN, middel is CMP with Gaussian prior and right is normal CMP.}
\label{exp1fig}
\vskip -0.1 in 
\end{figure}

We visualize the generated trajectories in Figure \ref{exp1fig} as qualitative analysis. As we can see, compared to the single deterministic prediction made by NN, CMP can generate more diverse and robust trajectories for a specified task. Quantitative results are shown in Table \ref{exp1table}. We observe that the estimated CLLs of CMP framework outperforms the baseline NN. Here we also compare an alternative of CMP, which samples latent variables simply from an isotropic multivariate Gaussian prior instead of the true latent distribution $p_{\boldsymbol{\theta}}$ during both training and inference. According to the results, the origin CMP achieves better performance than the alternative in most of the experiments, which shows that the latent variables are indeed conditional correlated with context and task.

\subsection{Ablation Analysis on Real Robot}
\textbf{Experiment Setup}~~~We use a 7-DOF Baxter robot as the main platform. Ten objects are used from YCB object set. The size of working area is 90cm$\times$90cm with a blue background. A Logitech C170 camera is placed over the top of the working scene to capture the context information. The entire system runs on ROS \cite{ROS} and the concrete motion in joint space is computed via inverse kinematics. For each experiments group, we train the model on a synthesized dataset with size 5000, then evaluate it on Baxter on 50 different task configurations. Each object was tested approximately 30 times.

\textbf{Evaluation Criteria}~~~While there might be multiple ways to approach a specified object, not all the trajectories are feasible for execution. Since we use inverse kinematics to compute the motion in joint space, some trajectories may not have valid corresponding movements(which we called \textit{unfeasible}). As a result, it is natural to separate the robot executions into three different categories:
\begin{enumerate}
    \item Success, where the robot successfully executes a motion and finish the specified task.
    \item Failed due to the unfeasible trajectory.
    \item Failed due to the wrong trajectory.
\end{enumerate}

\textbf{Performance Evaluation}~~~We set up three groups of experiments. In the first group, both Spatial Arg-Softmax(in CMPG) and LN is removed. Without target constraints provided by LN, MCS will work as a trajectory solver. In the second group, only LN is removed, while MCS will still remain as a trajectory solver. For the last group, we will use a complete CMP with a dynamic control system. All the other configurations will be the same among these groups besides the model.  

Table \ref{exp2table} shows the results of our ablations and an NN baseline. A full table of evaluation on each object is available in Appendix. There are several interesting facts in these results. First, we can see that the Spatial Arg-Softmax operation in CMPG is extremely important since it yields a significant improvement. Secondly, CMP $-$ LN is exactly the third model in Table \ref{exp1table}. Compared to the NN baseline, the later one is more likely to generate unfeasible trajectories, which shows that the proposed probabilistic model is more appropriate for trajectory learning. To look at the full table in Appendix, we can also find that the results between objects are not remarkably different. This means our model is successfully trained to focus more on object position instead of other irrelevant information.

At last, the combination of a dynamic control system can also yield improvements in the results as it provides the guarantee of reaching the goal. However, in our experiments, we find that the smoothness of weight $\mathbf{w}$ when learning the force space trajectory is not as good as position trajectory, which makes the model hard to converge when training CMP, and causes some unfeasible or even wrong trajectories. How to balance this trade-off could be worth some further research. 

\begin{table}
\centering
\begin{tabular}{l||c|c|c} 
\hline
Evaluation     & Success & Failure~(unfeasible) & Failure~(wrong)  \\ 
\hline
NN(baseline) & 39.7\%  & \textbf{14.0\%}  & 46.3\%  \\ 
CMP $-$ $\text{SAS}^*$ $-$ LN & 24.3\%  & 12.3\%  & \textbf{63.4\%}  \\ 
CMP $-$ LN       & 56.7\%  & 2.3\%  & 41.0\%  \\ 
CMP            & \textbf{75.7\%} & 13.0\%  & 11.3\%  \\
\hline
\multicolumn{4}{l}{* denotes Spatial Arg-Softmax}                
\end{tabular} \\
\vskip 0.1 in
   \caption{Results of baselines and ablations of proposed method.}
\label{exp2table}
\vskip -0.3 in 
\end{table}


\section{Related Work} 
Here we will review some previous work about movement primitives and motion planning. 

\textbf{Probabilistic Movement Primitives}~~~In this work~\cite{paraschos2013probabilistic}, the authors propose a probabilistic representation of trajectory. As it reduces the dimension of the trajectory with arbitrary length into a weight vector of basis function, it parameterizes the trajectory distribution by the distribution of weight and a fixed variance. The authors also show how to satisfy some traditional trajectory constraints including via point, target point and coupling by introducing equality constraints when solving the probabilistic model. Our probabilistic representation of trajectory was basically inherited from this work, however, compared to this work, we extend it to represent the trajectory in force space, which makes it possible to combine the probabilistic model with a dynamic control system. Besides, our CMP framework can learn from the contextual information and task at the same time, while~\cite{paraschos2013probabilistic} can only learn from the trajectory. Thus our model is more suitable for motion planning on real-world tasks. 

\textbf{Movement Primitive with Conditions}~~~A recent research work \cite{condtionMP} introduces a method for learning and retrieving movement primitives with condition input. The main idea of this work is measuring the distance between queries and movement primitives in a pre-built library. Once the metric is learned, we can select a subset of movement primitives with an input query based on the nearest principle, then output the average of the subset. It highly depends on the library of movement primitives, and the inner structure of the trajectory and condition remains unexplored. In contrast, our CMP framework models the conditional distribution of trajectory over contextual information and task. And with the help of the deep generative model, we can directly learn and inference with the complex posterior without any linear assumption \cite{usepromp} and achieve better generalization. 

\textbf{CNN based Movement Primitives}~~~In \cite{CNNDMP}, researchers try to use a forward neural network to learn the parameters of movement primitives directly from condition input(image of working scene). This model is exactly the same as the NN baseline we used in experiments. As we discussed in Section \ref{Probabilistic Trajectory Representation}, a trajectory is inherently to be stochastic instead of deterministic. Thus we believe that a stochastic model may be more appropriate for learning and inference the trajectory, and will lead to a better performance. The results of our experiments also support our intuition.

\section{Conclusion} 
In this work, we presented CMP, a novel framework aiming at a challenging problem in robotics: generate feasible movements on a specified task and corresponding contextual information. Experiments on both simulation and real-world shows that our approach can learn and inference movement effectively and stably. There are a lot of exciting directions for future work. One of them could be combining CMP with other sequential models to make long-term movement inference. Most importantly, we hope to apply our method to more challenging robotic tasks and explore its potential towards a general robotics trajectory and skill learning system.    


\bibliographystyle{plainnat}
\bibliography{nips_2018.bib}

\newpage

\end{document}